\documentclass[11pt]{article}

\usepackage{graphicx}
\usepackage{hyperref} 

\begin{document}
\date{22 September 2020}

\title{%
Multi-threaded Memory Efficient Crossover in C++ \\
for Generational Genetic Programming
}

\author{
\href{http://www.cs.ucl.ac.uk/staff/W.Langdon/}
{W. B. Langdon}
}

\maketitle

\pagestyle{plain}
\thispagestyle{empty}

\begin{abstract}
\noindent
  C++ code snippets from a multi-core parallel
  memory-efficient crossover for genetic programming are given.
  They may be adapted for separate generation evolutionary algorithms
  where large chromosomes or small RAM require no more than 
  $M+2\times\rm{nthreads}$
  simultaneously active individuals. 
\end{abstract}

\begin{figure}[!h]
\centerline{\includegraphics[scale=1]{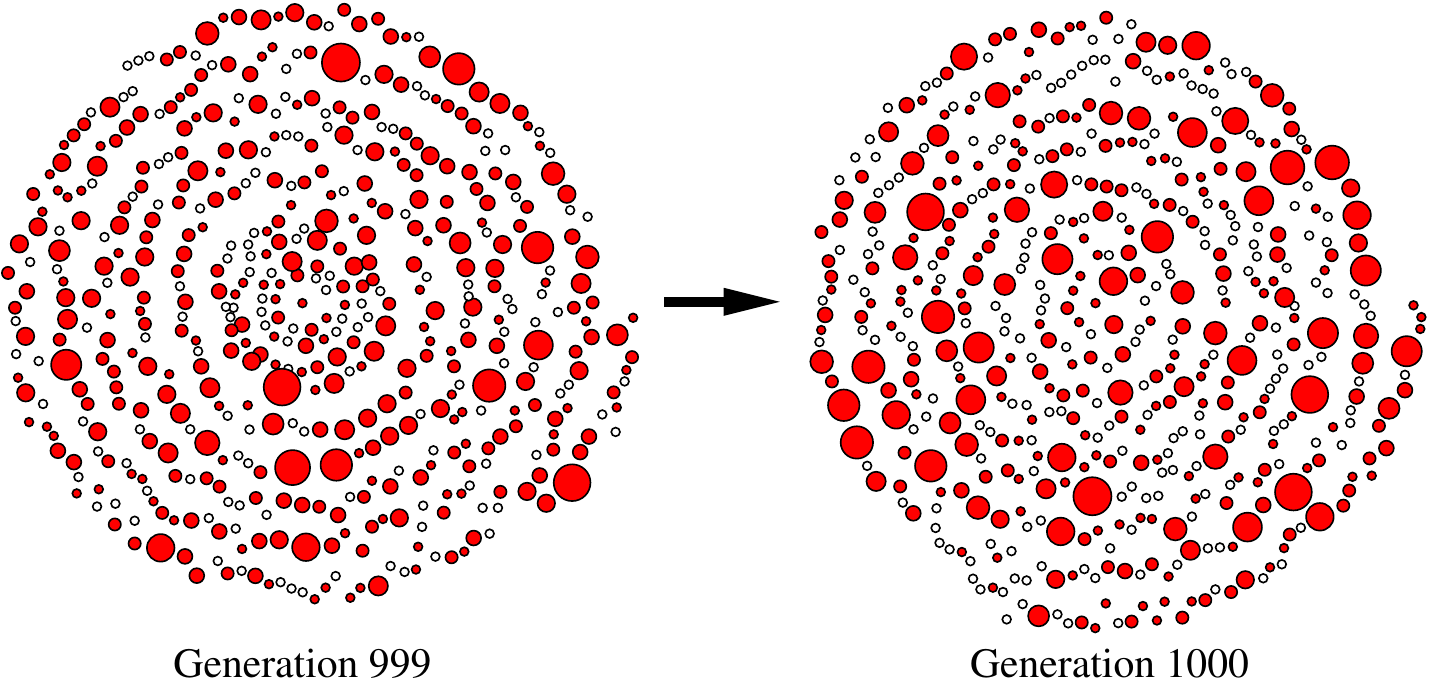}}
\caption{In generational schemes 
(e.g.~\mbox{$\mu , \mu$}) 
the current
population (on the left) 
is completely
replaced by the next population
(on the right). 
The area of red dots is proportional to the number of children
it will be a parent of in the next generation. 
Small white dots are infertile.
We describe how to avoid simultaneously storing both populations.
\label{fig:generational}
}
\end{figure}

\section{Background}
\begin{figure}
\centerline{\includegraphics[scale=1]{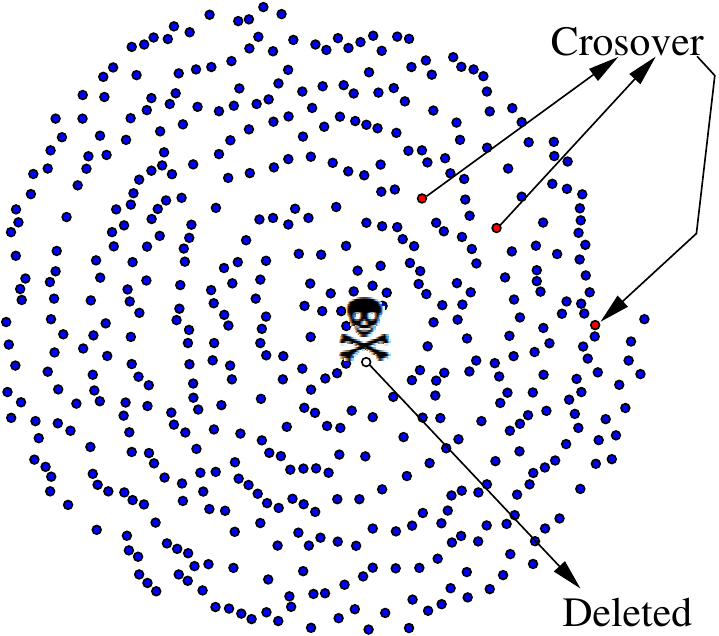}}
\caption{In steady state
Evolutionary Algorithms
(e.g.~\mbox{$\mu + 1$}) 
parents are selected from the current population,
their offspring created and inserted into the current population (red).
To keep the population size constant, 
the same number of individuals (white) are removed from the population.
Many strategies are available for selecting parents and 
for selecting who to remove from the population.
\label{fig:steady_state}
}
\end{figure}

It has been known for a long time that in conventional two parent genetic programming
\cite{koza:book,banzhaf:1997:book,poli08:fieldguide}
only $M+2$ individuals are required for non-overlapping generations of $M$~trees
\cite[pages 1044-1045]{koza:gp3}.
Indeed this is true of other forms of GP
\cite{nordin:thesis,
oneill:book,%
Miller:CGP}
and
generational%
\footnote{Steady state GA
\cite{srgs}, 
GPs \cite{singleton:byte},
and ($\mu + \lambda$)-ES Evolution Strategies
use overlapping generations,
Figure~\ref{fig:steady_state}.}
Genetic~Algorithms and Evolutionary Computation in general.
Nonetheless the widespread availability of large RAM computers and compact
coding of individual chromosomes seems to have led to the $M+2$~limit being
forgotten and the widespread use of 
inefficient computer implementations,
e.g.~\cite{langdon:2005:NC},
with separate new and old populations, each requiring storage for $M$ individuals.
Recent work with enormous trees
\cite{Langdon:2017:GECCO,Langdon:2019:alife}
or large populations~\cite{langdon:2008:SC} 
has prompted renewed interest in ensuring that 
generational implementations are as efficient as steady state implementations.
(For example, 
see email discussion on 
\href{https://groups.io/g/GeneticProgramming}
{GeneticProgramming@groups.io}
5--7 September 2020.)
 
\section{Minimal Memory Generational Algorithms}
\label{sec:alg}

Koza et al.~\cite[pages 1044-1045]{koza:gp3}
divide the current population into four classes,
according to how many children they have:
0, 1, 2, more than~2.
The new population is created in this priority order
(0, 1, 2, 2+).
Our algorithm is slightly simpler.
Since GPquick~\cite{singleton:byte}'s crossover creates one child (rather than two),
this allows
the ``2'' and ``more than 2'' classes to be combined into a class 
``two or more'' (2+).
We then follow \cite[pages 1044-1045]{koza:gp3},
except 
(to allow crossovers in parallel)
we allocate up to two free slots per thread
(rather than an extra two).

At the start of the new generation,
the parent population are assigned into classes 0, 1 and 2+.
All the parents without children (class~0) are deleted.
The children to be created are assigned into class 1 or~2+,
according to the minimum class of their (two) parents.
I.e.~if either parent is in class~1,
the new child is placed in class~1, otherwise class~2+.
With rapid parallel fitness evaluation,
the cost of crossover can be significant,
therefore both crossover and fitness evaluation are done by parallel threads~%
\cite{langdon:2019:gpquick}.
This means the remaining operations are also done in parallel. 

Creation and testing of new individuals then starts with
the class~1 individuals.
Each time a new individual has been created,
it is removed from the data structures for both its parents.
If it has a class~1 parent,
removing it will mean that that parent now
no longer has any children to be processed
and so is deleted.
Note this can be done before starting fitness evaluation of the child.  
Removing it from a class~2+ parent,
may mean the parent still has two or more children to be processed,
so it stays in class~2+
or it may now only have one child to be created,
in which case the child is moved to class~1.
Class~1 children (i.e.~with one or more class~1 parent)
are created before those 
with two class~2+ parents.

The amount of memory used in a particular 
multi-threaded run 
is between
$M+1$ and
$M+2\times\rm{nthreads}$,
and depends on the exact order of operations.
The assignment of individuals to threads
and the scheduling of those threads
typically varies, 
and so there can be variations in memory usage
between otherwise identical runs.

\pagebreak[4]
\section{C++ pthreads Implementation}

Our implementation is aimed at 
improving memory efficiency of 
extended runs (weeks)
with small populations 500--4000
with tournament selection~(7),
crossover
and trees of a billion nodes,
on multi-core parallel compute servers with modest RAM memory ($<1$TB). 
Thus the implementation must support parallel evaluation.
Secondly,
although our implementation should be widely applicable,
this environment lead to various design choices.

\subsection{Reusing expr Buffers}
\label{sec:expr}

In extended runs,
where GP trees are allowed to bloat 
\cite{Langdon:1997:bloatWSC2,%
poli08:fieldguide}
memory consumption is dominated by the space required to store the trees.
Since we are starting from a mature C++ code base
\cite{langdon:2019:gpquick},
we only optimise the {\tt expr} buffers used to store the trees.
That is,
although we could minimise the number of active 
individuals ({\tt chrome}) simultaneously in use,
for simplicity and minimising impact on the existing code,
we retain separate {\tt pop} and {\tt newpop} data structures
and concentrate upon {\tt expr}.
(The {\tt chrome} data structures hold accounting information and,
in our use case,
are tiny compared to the trees.)

Originally GPquick
allocated each {\tt chrome} and {\tt expr} buffer
as it was created and removed it when the {\tt chrome} was deleted.
Although GNU C++ fully supports multi-threaded 
dynamic memory allocation and de-allocation,
in practice we have found with multiple active threads this comes with
a high overhead.

New routine {\tt init\_exprs}
(see Appendix~\ref{sec:code})
creates and initialises an array of $M+2\times\rm{nthreads}$
pointers to hold addresses for {\tt expr} buffers
and a free {\tt chain} linking them all.
In a multi-threaded context,
these must be protected by a synchronisation lock. 
{\tt get\_expr} gives the next free {\tt expr} buffer 
to its calling {\tt chrome}.
If the chosen {\tt expr} buffer has not yet been allocated 
{\tt get\_expr} creates it.
(Note all {\tt expr} buffer are fixed length of {\tt pMaxExpr} bytes.)
In order to allow {\tt free\_expr}
to return the {\tt expr} buffer to the free {\tt chain} without searching,
{\tt get\_expr} 
also records which {\tt expr} buffer {\tt chrome} has been given 
in the new {\tt chrome} field {\tt expr\_id}.
Note once an {\tt expr} buffer has been allocated by C++ from the heap
it is never returned to the heap.
Instead {\tt get\_expr} and {\tt free\_expr} allow it to be reused
in each generation.

\subsection{Few Children per Parent}

In many evolutionary systems the populations tend to converge so that the breeding population 
(i.e.\ the existing individuals who will give rise to children)
is a sizable fraction (e.g.~30\%)
of the whole population.
Meaning, typically each only has a few children,
(e.g.\ on average about five).
Therefore the {\tt children} data structure uses simple linear arrays 
to hold the identifiers of the children of each parent
(Appendix~\ref{sec:children}).
At first sight this may seem inefficient.

Short linear scans along a simple vector to add/delete may be faster
than more complex data structures which avoid search.
Although we have not done this,
the target hardware supports AVX vector operations,
which can perform operations on up to 16 (or even 32) identifiers in one go.

Allocating and initialising each of the {\tt children} arrays
is eased by knowing before each generation starts
how many children each parent will have ({\tt num\_children}).
Thus each linear array can be easily created of the correct size.

As mentioned in the previous section,
in practice new/delete operations to allocate/remove 
heap data structures may be expensive when done with multiple active threads.
(Which is another reason for avoiding some alternative dynamic data structures.)
Therefore all {\tt children} arrays are pre-allocated
and initialised
in the default single master thread before any of the new population
are constructed by crossover
(Appendix~\ref{sec:generation_fitness}).
Similarly they are all de-allocated after the fitness' 
of the new generation has been evaluated.
Although the {\tt children} arrays are different sizes,
they are always small, 
and we have not noticed any problem with heap fragmentation.

\subsection{Integration with existing pthreads code}
\subsubsection{Additional master thread code}
In generational mode the existing
routine {\tt generation\_fitness}
(Appendix~\ref{sec:generation_fitness})
is used to sort parents and to be created children into 
class~0, 1 and 2+
(as described in Section~\ref{sec:alg}).
After deleting all parents without children (class~0),
{\tt generation\_fitness} creates {\tt nthreads}.
After the threads have all have finished,
it removes the {\tt children} arrays.
Notice the {\tt children} arrays are allocated from the heap
and returned to it, 
by conventional (non-parallel) code.

As before,
{\tt generation\_fitness} also removes all the parent {\tt chrome}
and switches to {\tt newpop}.
As mentioned in Section~\ref{sec:expr},
the {\tt chrome} data structures are tiny compared 
to the space needed to store their trees (i.e.~{\tt expr}).
Therefore, to limit impact on the existing code,
we only minimise the number of {\tt expr} buffers.

\subsubsection{Additions to the parallel crossover and fitness evaluations code}

The existing {\tt thread\_fitness} function 
(Appendix~\ref{sec:thread_fitness})
is executed separately
by each of {\tt nthreads}.
It consists of a loop,
which takes the next unevaluated new child,
created it by crossover,
and evaluates its fitness.
It then goes to the top of the loop to find the next 
member of the new generation to process.
It continues looping until it or the other threads
have processed everyone.

The principle changes are: 
\begin{itemize}
\item
Use {\tt chainhd1} and {\tt chainhd2}
to process the new population in class priority order
(rather than just numerical order).

\item
Use {\tt get\_expr} to allocate space for the new tree
(Section~\ref{sec:expr} and Appendix~\ref{sec:get_expr}).

\item
Use functions {\tt remchild} and {\tt move21}
to update class~0, 1 and~2+ information
(Section~\ref{sec:alg} Appendix~\ref{sec:remchild}).

\begin{itemize}
\item
{\tt remchild} is used
to remove the new individual from the {\tt children}
arrays used by its two parents ({\tt Mum} and {\tt Dad})
after it has been created by crossover
(but potentially before fitness evaluation).
Note due to multi threading
the classes may not be the same as when the child was selected
to be processed by the current thread.
(Even with a single thread
this can also be true of the second call to {\tt remchild}.)
Similarly due to potential self-crossover (i.e.\ overlapping parents),
{\tt remchild} must only remove one instance from the {\tt children} array.

Since {\tt remchild} is already going to process
most of each {\tt children} array,
it is a convenient place to see how many children remain
to be processed.
This is returned to {\tt thread\_fitness}.
Should there be only one child left for this parent,
its id is past via the {\tt last} argument
so that it is available to the {\tt move21} function.

\item
The {\tt move21} function (Section~\ref{sec:move21})
is actually simple,
but the listing is long as the implementation
contains extensive sanity checks of the class~1 and class~2+
chains.
{\tt move21} removes the indicated parent from the class~2+
chain
and then places it at
the head of the class~1 chain.
Since the parent can be arbitrary it must be removed from the 
class~2+ chain cleanly.
This is implemented using a bidirectional chain 
(i.e. with both {\tt forw} and {\tt back} pointers).
In contrast it can be inserted at the start of the class~1 chain 
(via {\tt chainhd1}, and so becomes the next item to process).
Thus the class~1 chain is not bidirectional
and its {\tt back} pointers are not maintained.

As mentioned above, {\tt move21} must deal with cases
where the child has already been removed from class~2+.
{\tt move21} uses
the array {\tt status} (values 0, 1 and 2, meaning class~0, 1 or 2+)
to avoid scanning the class~2+ chain.
(Also a check is needed that the child is not the one 
being processed. 
It is convenient to place that check in {\tt move21}.) 
\end{itemize}
\end{itemize}

\begin{figure}
\centerline{\includegraphics{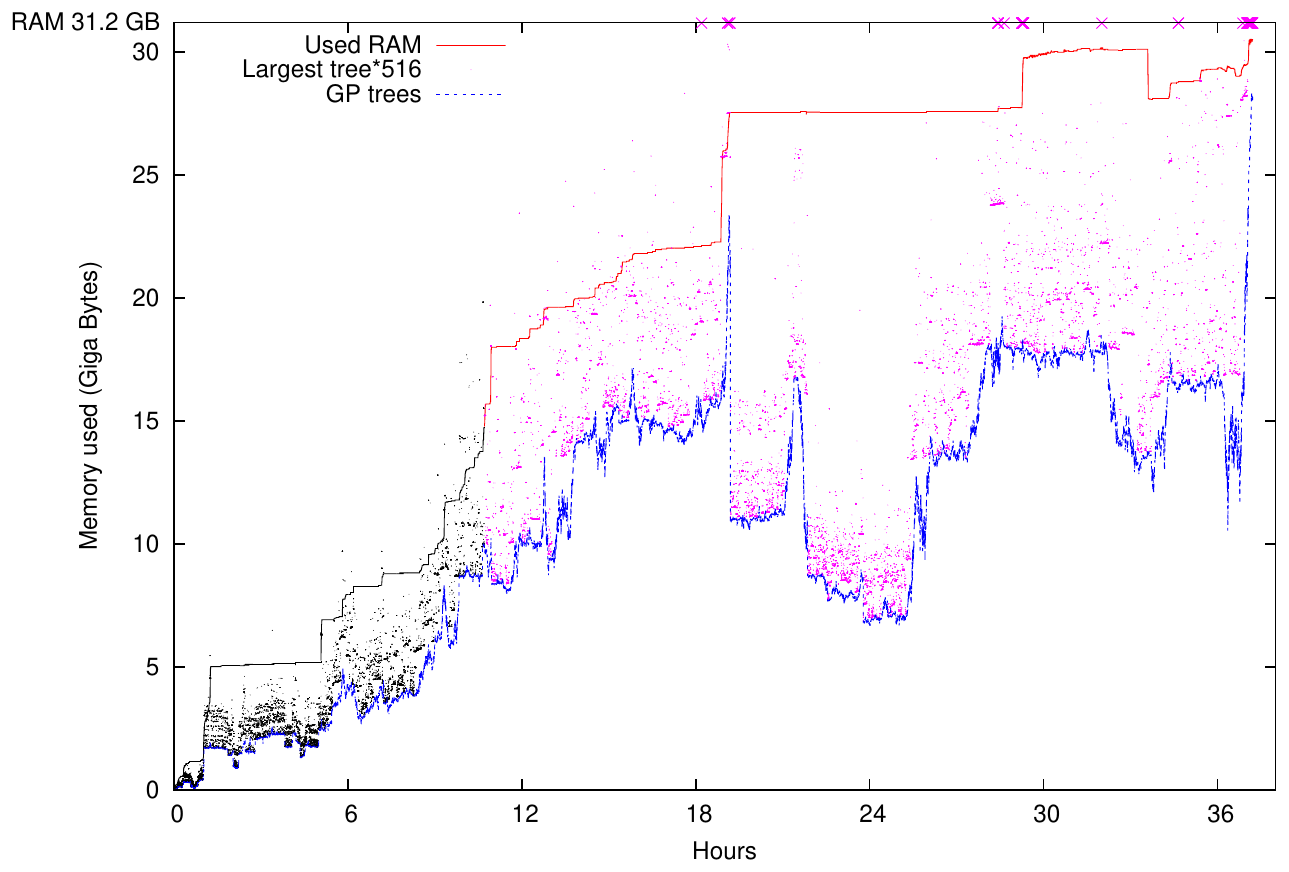}} 
\vspace*{-2ex}
\caption{\label{fig:sigevolution}
Example generational GP run on 8 core i7-4790 3.60GHz CPU 
with 31.2GB of memory.
The new efficient handling of memory
enables the run to continue 3.5 times as long as would have been possible previously
(shown in black).
Notice with the earlier limit,
the population bloats, 
wheres the new code 
(colour, after 11~hours)
shows GP gives more interesting dynamics if able to run for longer.
The run is terminated after generation 71\,248
when GP occupies 97.8\% of system memory
having processed the equivalent of
$2.61\,10^{16}$ GP opcodes in
37~hours 13~minutes
(195~Giga~GPop/s,
cf.~%
\mbox{\protect\cite[Table~7]{langdon:2015:hbgpa}},
\mbox{\protect\cite[Table~1]{langdon:2019:gpquick}}).
Notice although GPquick uses a fixed number (516) of fixed length
buffers 
($3.2\ 10^{8}$ bytes)
to hold the population,
the Unix process memory (solid red line) tracks ahead of the 
memory used to store the trees
and is seldom reduced when the trees become smaller 
(lower, dashed blue, line).
Note size of largest tree 
(purple dots or crosses~$\times$)
is rescaled by 516 to put on same scale as average tree
(dashed blue line).
When not gathering details statistics,
all but
0.14\% of run time is used by crossover and fitness evaluation
(i.e.\ in the 8~parallel threads).
However,
despite having 500 trees to evaluate with 8 threads,
variation in fitness times
leads to fast threads being idle waiting on the longest duration thread.
In this run this leads to losing
91.9\%
of a core.
I.e.~on average our generational GPquick effectively used only 7.08 cores in this run.
}
\end{figure}

\subsection*{Acknowledgement}

I would like to thank 
\href{http://www.dcs.bbk.ac.uk/~tom/}
{Thomas H. Westerdale},
\href{http://ernesto.dei.uc.pt}
{Ernesto Costa},  
Craig Shelden,  
David Oranchak, 
\href{https://www.napier.ac.uk/people/kevin-sim}
{Kevin Sim},      
\href{http://www.genetic-programming.com/johnkoza.html}
{John Koza},      
Simon Waite     
and
\href{http://research.fh-ooe.at/de/staff/15924}
{Gabriel Kronberger}. 

\bibliographystyle{abbrvurl1}
\bibliography{gp-bibliography,references}

\newpage
\appendix
\section{GPquick C++ Code Snippets. chrome.cxx Revision: 1.262}
\label{sec:code}

Standard C++ conditional compilation
\verb'#ifndef NDEBUG'
and \verb'assert( )'
are used only during development 
and are disabled when compiled with \mbox{\tt -O3 -DNDEBUG}.
\newline
\url{http://www.cs.ucl.ac.uk/staff/W.Langdon/ftp/gp-code/efficient_memory/}

\subsection{expr buffers}

\begin{verbatim}
void Pop::init_exprs(const int nexpr){
  assert(thispop==NULL);
  thispop = this;
  assert(chain==NULL);
  const int n = nexpr+1;
  chain   = new     int[n]; //leave 0 unused
  exprptr = new nodeptr[n]; //leave 0 unused
  memset(exprptr,0,n*sizeof(node*));
  chain[0] = 0;             //be tidy
  for(int i=1;i<nexpr;i++) chain[i] = i+1;
  chain[nexpr] = 0;         //end of chain
  chainmax  = nexpr;        //for debug
  chainhead = 1;
}
\end{verbatim}

\begin{verbatim}
class Pop ...//NB protect with external locks
	int used      =  0; //only for performance monitoring
	int max_used  =  0; //only for performance monitoring
\end{verbatim}

\label{sec:get_expr}
\begin{verbatim}
void Pop::get_expr(Chrome* chrome){
  assert(chainhead<=chainmax);
  if(chainhead<=0) {
    cout<<"ERROR ran out of expr buffers "
        <<popsize<<" "<<chainmax<<" "<<chainhead
        <<" pThreads: "<<params->params[pThreads]<<endl;
    exit(1);
  }
  if(exprptr[chainhead]==0){
    exprptr[chainhead] = new node[params->params[pMaxExpr]];
  }
  chrome->expr = exprptr[chainhead];
  chrome->expr_id = chainhead;
  chainhead = chain[chainhead];
  assert(chainhead==0 || (chainhead>=1 && chainhead<=chainmax));
  used++;
  if(used>max_used){
    max_used = used;
    const int nthreads = params->params[pThreads];
    const int nexpr = popsize + 2*( (nthreads==0)? 1 : nthreads);
    if(nthreads && max_used > nexpr) cout<<"ERROR max_used increased to "<<max_used<<endl;
  }
}
\end{verbatim}

\pagebreak[4]
\begin{verbatim}
void Pop::free_expr(Chrome* chrome){
  const int id = chrome->expr_id;
  if(id==0) return; //already freed
  assert(id>=1 && id<=chainmax);
  assert(exprptr[id]!=0);
  assert(chainhead==0 || (chainhead>=1 && chainhead<=chainmax));

  chain[id] = chainhead;
  chainhead = id;
  assert(chainhead==0 || (chainhead>=1 && chainhead<=chainmax));
  used--;
  assert(used>=0);
  chrome->expr_id = 0; //allow free_expr to be called more than once
}
\end{verbatim}

\subsection{Tracking Children to be Created by Crossover
and Freeing their Parents}
\label{sec:children}

Note whilst {\tt expr} buffers uses 0 to
denote an invalid value, e.g. end of chain,
the following code uses -1.

\begin{verbatim}
int* forw      = NULL;  //valid 0..popsize-1
int* back      = NULL;  //valid 0..popsize-1, only maintained for chain2
int** children = NULL;  //popsize NULL or pointer to num_children int, -1 or 0..popsize-1
int* status    = NULL;  //0, 1 or 2
int chainhd1 = -1; //valid 0..popsize-1
int chainhd2 = -1; //valid 0..popsize-1
\end{verbatim}

\subsubsection{\tt generation\_fitness}
\label{sec:generation_fitness}

Fragment of code in master thread used 
at the start of each generation (except the first random population) 
in function {\tt generation\_fitness}.
\begin{verbatim}
  if(DoCross) { //100% crossover on later generations
    if(forw      == NULL) forw      = new int[popsize];
    if(back      == NULL) back      = new int[popsize];
    if(status    == NULL) status    = new int[popsize];
    if(children  == NULL) children  = new int*[popsize];
#ifndef NDEBUG
    memset(forw,     0x7f,popsize*sizeof(int));
    memset(back,     0x7f,popsize*sizeof(int));
    memset(status,   0xff,popsize*sizeof(int));
#endif
    memset(children,0,popsize*sizeof(int));
    int last1 = -1;
    int last2 = -1;
    chainhd1  = -1;
    chainhd2  = -1;
    for(int s=0;s<popsize;s++) {
      assert(newpop[s]->mum.birth >= 0);
      assert(newpop[s]->dad.birth >= 0);
      const int Mum = newpop[s]->mum.birth % popsize;
      const int Dad = newpop[s]->dad.birth % popsize;
      const Chrome* mum = pop[Mum];
      const Chrome* dad = pop[Dad];
      assert(mum->num_children>0);
      assert(dad->num_children>0);
      if(mum->num_children==1 || dad->num_children==1) {
        append(s, last1, chainhd1);
        status[s] = 1;
      } else {
        append(s, last2, chainhd2);
        status[s] = 2;
      }
      addchild(Mum,mum->num_children,s);
      addchild(Dad,dad->num_children,s);
    }//endfor all newpop

    for(int s=0;s<popsize;s++) {
      if(pop[s]->num_children==0) {delete pop[s]; pop[s] = NULL;}
    }
  }//endif DoCross init
\end{verbatim}

\subsubsection{Functions used by {\tt generation\_fitness}}

\begin{verbatim}
inline void append(const int s, int& last, int& chainhd) {
  const int popsize = thispop->popsize;
  assert(s>=0 && s < popsize);
  if(last != -1) {
    assert(forw[last] == -1);         forw[last] = s; }
  assert(  forw[s]    == 0x7f7f7f7f); forw[s]    = -1; 
  assert(  back[s]    == 0x7f7f7f7f); back[s]    = last;
  last = s;
  if(chainhd == -1) chainhd = s;
}
\end{verbatim}

\begin{verbatim}
inline void addchild(const int P,const int num_children,const int s) {
  //use linear search assuming num_children is small
  const int popsize = thispop->popsize;
  assert(P>=0 && P < popsize);
  assert(s>=0 && s < popsize);
  //if(!(num_children > 0 && num_children < 20)) cout<<"num_children "<<num_children<<endl;
  if(children[P]==NULL) {
    children[P] = new int[num_children];
    memset(children[P],0xff,num_children*sizeof(int));
  }
  int i=0; 
  for(;i<num_children;i++){
    if(children[P][i] == -1){children[P][i]=s;break;}
  }
  assert(i < num_children);
  assert(children[P][i]==s);
}
\end{verbatim}

\subsubsection{Multi-threaded code, i.e.\  {\tt thread\_fitness}}
\label{sec:thread_fitness}

Fragments of multi-threaded code used 
during crossover and fitness evaluation
in each generation (except the first) 
in function {\tt thread\_fitness}.

Allocate new child {\tt i} for the thread to work on.
Choose {\tt i} from class~1 if any,
otherwise from class~2+.
If both classes are empty, there is no more work,
so stop the thread. 
\vspace*{-1ex}
\begin{verbatim}
    Chrome* newguy;
    {const int e = pthread_mutex_lock(&mutex);   assert(e==0);}
    if(DoCross) { //100% crossover on later generations
    int i = -1;
    if(       chainhd1 != -1) {
      i = chainhd1;
      chainhd1 = forw[i];
    } else if(chainhd2 != -1) {
      i = chainhd2;
      chainhd2 = forw[i];
    }
    if(i == -1) {
      {const int e = pthread_mutex_unlock(&mutex); assert(e==0);}
      break;
    }
    assert(status[i]==1 || status[i]==2);
    status[i] = 0; //make sure no other thread tries to process i
    newguy = thispop->newpop[i];
    newguy->thread = my_id;
    assert(newguy->expr == NULL);
    thispop->get_expr(newguy);
    {const int e = pthread_mutex_unlock(&mutex); assert(e==0);}
\end{verbatim}

\noindent
Do crossover. Reading from Mum and Dad writing new child into {\tt newguy}.
 
After crossover, 
unless doing incremental fitness,
now ok to remove {\tt newguy}
from both its parents.
If either parent was in class~2+ but now only has one child left
to be processed,
the parent is moved to class~1.
If either parent now has no children left to process,
it is deleted.
Note freeing {\tt mutex} to allow crossovers to be done in parallel
means other threads may have already processed other children 
with the same parents,
so, for example, 
a parent may no longer be in class~2+ when {\tt move21} is called
to move it from~2+ to 1. 
\vspace*{-1ex}
\begin{verbatim}
    {const int e = pthread_mutex_lock(&mutex);   assert(e==0);}
    int last1, last2;
    const int nchild1 = remchild(Mum,mum->num_children,i,last1);
    const int nchild2 = remchild(Dad,dad->num_children,i,last2);

    if(nchild1 == 1) move21(i,last1);
    if(nchild2 == 1) move21(i,last2);
    if(nchild1 == 0) {
      mum->num_children = 0;
      thispop->free_expr(mum);    }
    if(nchild2 == 0) {
      dad->num_children = 0;
      thispop->free_expr(dad);    }
    {const int e = pthread_mutex_unlock(&mutex); assert(e==0);}
\end{verbatim}

\noindent
Do fitness evaluation for {\tt newguy}
and then loop to top of 
function {\tt thread\_fitness}
to see if there is any more work to be done.

\subsubsection{Multi-threaded Functions used by {\tt thread\_fitness}}
\label{sec:remchild}

\begin{verbatim}
inline int remchild(const int P,const int num_children,const int s, int& last){
  //use linear search assuming num_children is small
  const int popsize = thispop->popsize;
  assert(P>=0 && P < popsize);
  assert(s>=0 && s < popsize);
  int S = s;
  int nchild = 0;
  int ok = 0;
  for(int i=0;i<num_children;i++){//remove only one child
    assert(children[P][i] >= -1 && children[P][i] < popsize);
    if(children[P][i] ==  S) {children[P][i] = -1; ok++; S=-2;}
    if(children[P][i] != -1) {last = children[P][i]; nchild++;}
  }
  assert(ok==1);
  if(nchild != 1) last -1; //be tidy
  return nchild;
}
\end{verbatim}

\label{sec:move21}
\begin{verbatim}
inline void move21(const int active, const int s) {
  if(active == s)    return; //ignore
  if(status[s] != 2) return; //already processed
  //remove s from chain2
  status[s] = 1;
  const int b = back[s];
  const int f = forw[s];
  if(chainhd2 == s) chainhd2 = f;
  forw[s] = -1;
  back[s] = -1;
  if(b != -1) forw[b] = f;
  if(f != -1) back[f] = b;

  //Insert s at head of chain1
  forw[s] = chainhd1;
  chainhd1 = s;
}//end move21
\end{verbatim}

\end{document}